%% file: main.tex
\documentclass[runningheads]{llncs}

\usepackage[mobile]{eccv}

\usepackage{eccvabbrv}

\usepackage{graphicx}
\usepackage{booktabs}

\usepackage[accsupp]{axessibility}  

\usepackage{hyperref}

\usepackage{orcidlink}

\usepackage{multirow}

\begin{document}

\title{AR Forcing: Towards Long-Horizon Robot Navigation World Model} 


\author{Yifei Yang\inst{1} \and
Zehua Fan\inst{2} \and
Huan Li\inst{1} \and
Aoqi Wang\inst{1} \and
Lida Huang\inst{3} \and
Haibao Yu\inst{4} \and
Haiyan Liu\inst{5} \and
Xuanyao Mao\inst{5} \and
Jason Bao\inst{5} \and
Liang Xu\inst{5} \and
Bingchuan Sun\inst{5}$^\dagger$ \and
Yan Wang\inst{1}$^\dagger$}

\authorrunning{Y. Yang~Author et al.}

\institute{Institute for AI Industry Research, Tsinghua University \and 
Shanghai Jiao Tong University
\and
School of Safety Science, Tsinghua University
\and
The University of Hong Kong
\and
Lenovo, Beijing, China}

\maketitle
\begingroup
\renewcommand\thefootnote{}
\footnotetext{$^\dagger$ Corresponding authors.}
\endgroup
\begin{center}
    \centering
    \captionsetup{type=figure}
    \includegraphics[width=0.9\linewidth]{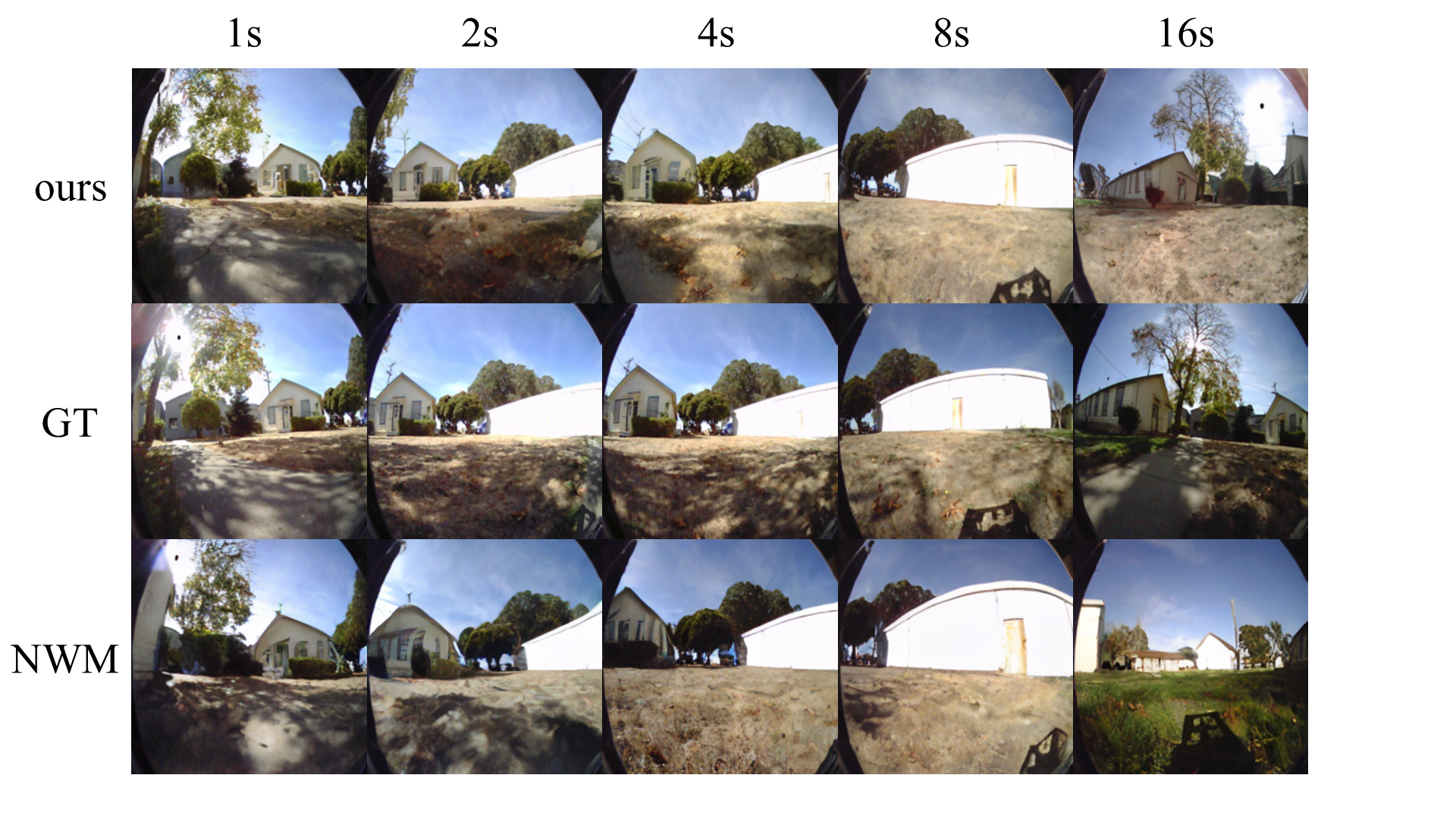}
    \captionof{figure}{\textbf{Qualitative comparison of long-horizon video generation.} We compare AR Forcing (ours) against the state-of-the-art NWM and the Ground Truth (GT) on the RECON dataset across a 16s rollout horizon. NWM drifts after 8s, while our method maintains high-fidelity structure and temporal consistency, closely matching GT after 16s.}
    \label{fig:top_view}
\end{center}

\begin{abstract}
  The diffusion based robot navigation world models are typically trained using parallel supervision, while autoregressive inference is employed during path planning. This results in a distribution shift between training and inference, which destabilizes the performance over long-horizon prediction. We propose AR Forcing, an autoregressive training strategy, which integrates the standard diffusion loss into the autoregressive training loop. At each step, the model uses its own predictions to update the context and optimize the single step noise prediction objective, thereby explicitly exposing the model to the inference state distribution during training. Our method does not require additional discriminators or distribution-matching losses, retains the original diffusion framework and sampler, and is easy to integrate. Experiments on multi-domain navigation datasets (RECON, SCAND, HuRoN, TartanDrive) show that compared with strong baselines, AR Forcing improved the consistency of generated images during long-horizon navigation and the accuracy of predicted trajectories, enhancing robustness of the model in complex known and unknown environments. We will release the code soon.
  \keywords{Diffusion Models \and World Models \and Autoregressive Training}
\end{abstract}

\section{Introduction}
\label{sec:intro}

Predictive world models~\cite{ha2018world} are a foundational component of modern robotics and embodied AI. A model that can accurately imagine future observations conditioned on actions enables planning without costly real world execution, supports counterfactual reasoning, and allows systems to evaluate alternative strategies at test time~\cite{wu2023daydreamer,micheli2022transformers,zhang2023storm}. As a result, world models have become central to tasks such as navigation, manipulation, and autonomous driving, where long-horizon prediction is crucial for downstream control~\cite{zhang2023copilot4d}.

Diffusion models have recently emerged as a powerful backbone for world modeling. Compared to earlier autoregressive or GAN based approaches, diffusion models offer strong generative fidelity, stable training dynamics, and the ability to represent multi-modal futures~\cite{tian2024visual,maze2023diffusion,sahoo2025diffusion}. By operating in a latent space and iteratively denoising, diffusion world models can synthesize high quality future frames and support rich visual prediction~\cite{blattmann2023align,yu2023video,yu2024efficient}. This has made them attractive for embodied tasks where perceptual realism is important and where downstream planners benefit from accurate simulated rollouts~\cite{zhang2023copilot4d,alonso2024diffusion}.

Despite these advantages, diffusion world models still face a critical limitation: long-horizon rollouts are often unstable. Short-term predictions can be plausible and sharp, but as prediction length increases, errors accumulate, leading to drift, temporal inconsistency, and eventually unrealistic or uninformative rollouts~\cite{alonso2024diffusion,huang2025self}. This instability directly affects planning quality: If the model's long-horizon predictions are unreliable, planning will select suboptimal or even unsafe actions.

A core reason for this instability is the train and test mismatch inherent to current training practices. In most implementations, diffusion world models are trained using parallel supervision: the model is conditioned on a fixed, ground-truth context and supervised to predict multiple future frames in parallel. However, inference is autoregressive, where the model feeds its own predictions back into the context step by step~\cite{huang2025self,chen2024diffusion}. The model is therefore trained under a context distribution that it never encounters at test time. This discrepancy induces \emph{exposure bias}: the model is optimized under $p_{\text{data}}(c)$, but must operate under $p_{\theta}(c)$, the distribution of contexts produced by its own rollouts.

This mismatch is a well-known issue in sequence modeling, and several solutions have been proposed in other domains, including explicit distribution-alignment objectives. For diffusion based sequence and video generation, recent work directly targets exposure bias by making training simulate inference time conditioning~\cite{huang2025self,chen2024diffusion}. For diffusion models more broadly, distillation, adversarial and consistency objectives are widely used to better match generation dynamics to desired behaviors or faster samplers~\cite{song2023consistency,sauer2024adversarial}.

From a practical perspective, this motivates a simpler question: Can we reduce the train and test gap by changing only the training schedule without changing the diffusion loss, the model architecture, or the sampler? Such a method would be easy to deploy in existing systems and would preserve the empirical benefits of diffusion training while explicitly addressing rollout mismatch.

In this work we answer this question affirmatively. We propose an autoregressive training strategy, AR Forcing, which is for diffusion world models. The method retains the standard diffusion objective but modifies the training loop to simulate inference time rollouts. At each step, the model denoises a noisy latent for the next frame, uses its own prediction to update the context, and continues to the next step. The loss remains the conventional single step diffusion objective, but the conditioning distribution now reflects the model's own rollout states.

This design has several advantages. First, it is simple and compatible: it plugs directly into existing diffusion training code and can be initialized from pretrained checkpoints. Second, it is efficient: it avoids full sampling based inner loops by using a single diffusion loss per step. Third, it is principled: by aligning the training context distribution with the inference distribution, it directly targets the source of exposure bias in long-horizon prediction.

We evaluate the approach on four multi-domain navigation datasets (RECON~\cite{shah2021rapid}, TartanDrive~\cite{triest2022tartandrive}, SCAND~\cite{karnan2022socially}, HuRoN~\cite{hirose2023sacson}) and compare against the standard parallel supervised diffusion baseline, NWM~\cite{bar2025navigation}. We find that AR Forcing is more stable, particularly at long-horizon prediction, where error accumulation is most severe. Importantly, the gains translate to improvements in closed-loop planning performance, as measured by trajectory and goal reaching metrics. Figure \ref{fig:top_view} compares the performance of our generated images with the baseline at different times. These results indicate that aligning training dynamics with inference dynamics is critical for long-horizon world modeling.

We summarize our main contributions as follows:
\begin{itemize}
    \item We identify the train--test mismatch inherent in diffusion world models as a key source of long-horizon instability. To address this, we propose a simple yet effective AR Forcing training schedule.
    \item We demonstrate that aligning the training and inference distributions can be achieved using the standard diffusion objective alone, eliminating the need for auxiliary components such as discriminators, specialized losses, or architectural modifications.
    \item We empirically validate the proposed method on navigation benchmarks, showing that it yields more stable rollouts and leads to improved downstream planning performance, particularly under long-horizon evaluation settings for known and unknown environments.
\end{itemize}

\section{Related Work}

\subsection{Diffusion World Models and Visual Prediction}
Diffusion models have become a dominant paradigm for high-fidelity image generation and are increasingly applied to video prediction and world modeling~\cite{blattmann2023align,yu2023video}. In embodied and navigation tasks, diffusion world models are trained to predict future observations conditioned on context frames and actions, including
discrete diffusion world modeling for autonomous driving~\cite{zhang2023copilot4d} and diffusion world models for interactive environment dreaming~\cite{alonso2024diffusion,valevski2024diffusion,yang2023learning}. Compared to pixel-level autoregressive predictors, diffusion models tend to be more stable to train and provide stronger perceptual quality, which is attractive for long-horizon rollout and planning~\cite{alonso2024diffusion,yang2023unisim,hu2023gaia,liu2025aligning}. However, most diffusion world models are still trained using parallel supervision: given a fixed, ground-truth context, the model predicts future frames independently under the standard diffusion objective~\cite{zhang2023copilot4d,bruce2024genie,wang2024driving,chen2024taming}. This training protocol differs from inference, where the model must roll out autoregressively using its own predictions.

\subsection{Train--Test Mismatch and Exposure Bias}
The gap between training on ground truth contexts and inference on model generated contexts is a long standing issue in sequential prediction.
In vision, language, and control, exposure bias leads to compounding errors and unstable long-horizon rollouts~\cite{ning2023elucidating,zhang2025anti}.
Recent diffusion sequence work explicitly targets this mismatch by changing the training or sampling paradigm to better support long rollouts and decision making, for example per-token noise levels with causal generation and training strategies that bridge rollout conditioning~\cite{chen2024diffusion,huang2025self,asada2025addressing,ruhe2024rolling,cui2025self}. Rolling Forcing ~\cite{liu2025rolling} uses a rolling-window joint denoising scheme to reduce error accumulation and conditions training on self-generated histories to mitigate exposure bias.
In diffusion settings, several methods propose distribution-matching objectives~\cite{yin2024one,yin2024improved,zhou2024score,geng2023one}, such as distillation and adversarial terms to accelerate sampling~\cite{sauer2024adversarial} or align generated distributions with teacher diffusion priors~\cite{song2023consistency, luo2023diff}.
Compared to Self Forcing approaches~\cite{huang2025self,cui2025self} that add explicit distribution‑matching losses (e.g., DMD, SID, GAN or discriminator‑based alignment), we do not introduce any extra objectives or networks; instead, we expose the model to its own rollout contexts while keeping the standard diffusion loss unchanged.

\subsection{World Models for Planning}
World models are widely used for model based planning, where candidate action sequences are rolled out in the learned model and scored by a cost function~\cite{hafner2023mastering,wu2023daydreamer}.
The quality of the planner depends not only on perceptual sharpness but also on long-horizon stability and trajectory consistency.
Recent systems leverage diffusion generative modeling for planning and trajectory synthesis in offline RL and control, including multi-task diffusion planning~\cite{chi2025diffusion},
hierarchical diffusion planning~\cite{chen2024simple}, latent action diffusion planning~\cite{venkatraman2023reasoning}, and model based diffusion for trajectory optimization~\cite{lee2025local}.
In navigation, diffusion trajectory generation has also been used to produce temporally coherent trajectory sequences to guide decision making~\cite{shah2022gnm,sridhar2024nomad,bar2025navigation}.
In this setting, even small rollout drift can lead to mis-ranking and poor decisions~\cite{ni2023metadiffuser,sun2023conformal}. Our work directly targets this issue by aligning the training distribution with the inference distribution, improving the reliability of rollouts and thereby improving downstream planning metrics.

\begin{figure}[t!]
    \centering
    \includegraphics[width=1.0\linewidth]{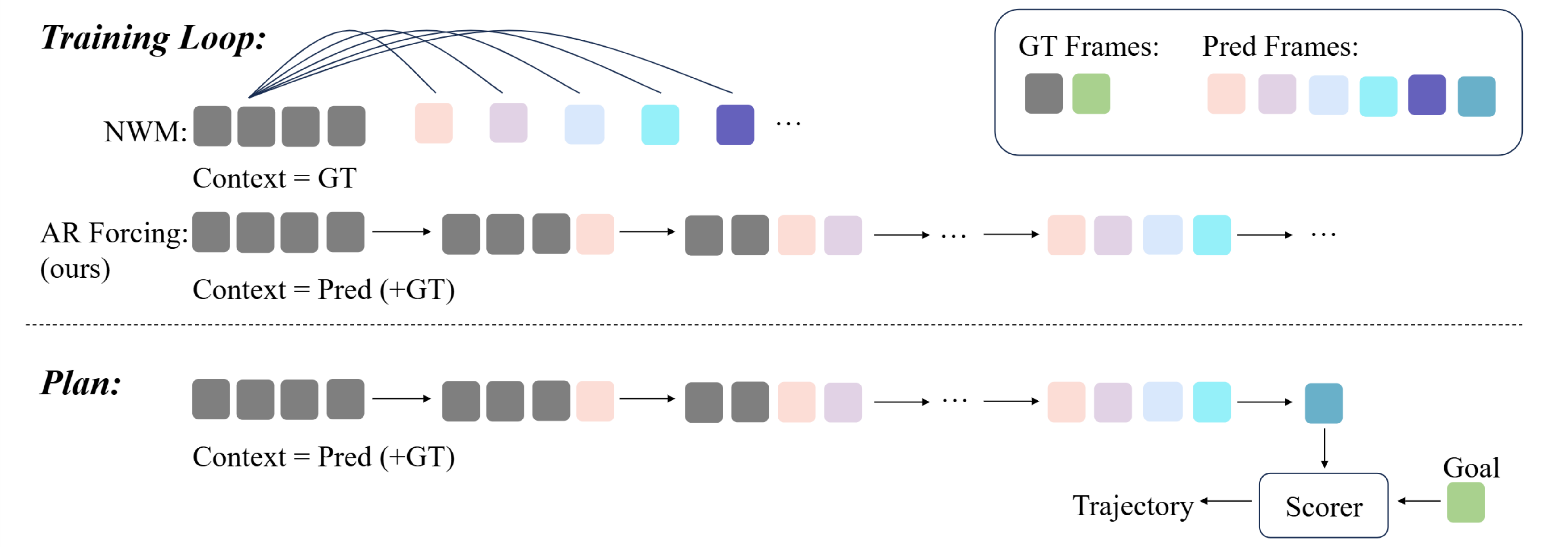}
    \caption{ \textbf{Architecture of AR Forcing.} In AR Forcing (ours), the model predicts the next frame, feeds its own prediction back into the context, and applies the same diffusion loss at each step. This aligns training with inference rollouts without extra losses or architectural changes, improving long‑horizon stability and planning robustness. NWM trains diffusion world models with parallel supervision: given a fixed ground‑truth context, it predicts multiple future frames under the standard diffusion objective. This creates a train–test mismatch, since inference rolls out autoregressively using model‑generated frames.}
    \label{fig:method_view}
\end{figure}

\section{Method}

\subsection{Preliminaries}
We study a conditional diffusion world model that predicts future visual observations conditioned on past context and action sequences. The model operates in the latent space of a fixed VAE~\cite{blattmann2023stable} for efficiency. Let $E(\cdot)$ and $D(\cdot)$ denote the VAE encoder and decoder~\cite{blattmann2023stable}, respectively. Given an image frame $I$, its latent representation is $x_0 = E(I)$, where $x_0 \in \mathbb{R}^{C \times H \times W}$. The decoder reconstructs pixels as $\hat{I} = D(x_0)$.

\subsubsection{Diffusion Process}
We adopt a standard DiT~\cite{peebles2023scalable} with $T_{\text{diff}}$ timesteps. The forward process adds Gaussian noise to a clean latent $x_0$ to obtain $x_t$. The denoiser $G_\theta$ predicts either the noise $\epsilon$ (epsilon-parameterization) or the clean latent $x_0$ ($x_0$-parameterization):
\begin{equation}
\hat{\epsilon} = G_\theta(x_t, t \mid c, y, r).
\end{equation}
Here $c$ is the latent context sequence, $y$ is the action conditioning, and $r$ is a relative-time embedding encoding the temporal distance of the prediction target. The canonical diffusion loss is
\begin{equation}
\mathcal{L}_{\text{diff}}(\theta) =
\mathbb{E}_{c \sim p_{\text{data}}}\,
\mathbb{E}_{x_0^*, y, r \sim p_{\text{data}}(\cdot \mid c)}\,
\mathbb{E}_{t,\epsilon}
\left[\left\|\epsilon - G_\theta(x_t,t \mid c, y, r)\right\|^2\right],
\label{eq:diffusion_loss}
\end{equation}
where $\theta$ is the model parameters; $x_0^*$ is the clean target latent; $t$ is the diffusion timestep;
$\epsilon\sim\mathcal{N}(0,I)$ is Gaussian noise; $x_t=q(x_t\mid x_0^*,t)$ is the noised latent; and
$G_\theta$ is the denoiser.
The loss is averaged over spatial dimensions and batch elements. When variance is predicted, the standard variational bound term is added, following common diffusion training practice~\cite{nichol2021improved}.

\subsubsection{World Model Setting}

Each training example consists of a context of past frames $\{I_1,\dots,I_K\}$, a sequence of future actions $\{a_1,\dots,a_T\}$ and corresponding future frames $\{I^*_{K+1},\dots,I^*_{K+T}\}$. We encode the context into latents $c_0 = [x_{0,1},\dots,x_{0,K}]$ and the future frames into targets $\{x^*_{0,1},\dots,x^*_{0,T}\}$. The model must generate a future latent trajectory conditioned on the context and actions. At inference time, predictions are generated autoregressively: the model's own predicted latents are appended to the context and used as input for the next step. This inference behavior motivates our training procedure in the next section.

\subsection{AR Forcing}
We now formalize the AR Forcing training objective and provide a detailed mathematical justification for why it better matches inference than parallel supervision.

Let a trajectory include latents $\{x_{0,1},\dots,x_{0,K+T}\}$ and actions $\{a_1,\dots,a_T\}$. The initial context is
\begin{equation}
c_0 = [x_{0,1},\dots,x_{0,K}] \in \mathbb{R}^{K \times C \times H \times W}.
\end{equation}
At step $i$, the model conditions on context $c_{i-1}$ and action $y_i$ (derived from $a_i$) to predict the future latent $x_{0,i}^*$ (the ground truth at that step).

We denote the data context distribution by $p_{\text{data}}(c_{i-1})$, induced by ground truth rollouts, and the model induced context distribution by $p_\theta(c_{i-1})$, induced by the model's own autoregressive predictions. These two distributions generally differ, and this discrepancy is the source of exposure bias.

\subsubsection{Context Update}
In AR Forcing, the context is updated with the model's own predictions at each step. We adopt the predicted-$x_0$ update because it is efficient and consistent with the diffusion objective. When the model is epsilon-parameterized, we reconstruct the clean latent via the standard inversion:
\begin{equation}
\hat{x}_0 = \frac{1}{\sqrt{\bar{\alpha}_t}}\left(x_t - \sqrt{1-\bar{\alpha}_t}\,\hat{\epsilon}\right),
\end{equation}
where $\bar{\alpha}_t$ is the cumulative product of alphas. The predicted latent $\hat{x}_0$ is then appended to the context:
\begin{equation}
c_i = [c_{i-1}[1:K], \hat{x}_{0,i}].
\end{equation}
If the model is $x_0$-parameterized, the output directly yields $\hat{x}_0$ and the same update applies. This design keeps the training loop lightweight: each step uses a single diffusion loss and a single context update, avoiding expensive inner denoising loops. We treat the updated context as a fixed input for the next step, which exposes the model to its own prediction distribution while maintaining stable optimization.

In practice, to maintain training stability and memory efficiency, we apply a \text{stop-gradient} operator on $\hat{x}_{0,i}$ before it is appended to the context $c_i$. This ensures that the gradient of the loss at step $i+1$ does not backpropagate through the entire diffusion sampling chain of the previous step, effectively treating the updated context as a ``sampled'' observation from the model's current belief. This decoupling prevents vanishing or exploding gradients across long sequences and significantly reduces the GPU memory footprint during training.

\subsubsection{AR Forcing Objective}
AR Forcing modifies only the conditioning distribution, not the diffusion loss. We roll out the model during training and update the context with predictions. 
The AR Forcing objective is therefore:
\begin{equation}
\mathcal{L}_{\text{AR}}(\theta) = \mathbb{E}_{c_0 \sim p_{\text{data}}} \,
 \mathbb{E}_{c_{i-1} \sim p_\theta(\cdot\mid c_0)} \,
 \mathbb{E}_{x^*_{0,i},y_i,r_i} \, \mathbb{E}_{t,\epsilon}
\left[\left\|\epsilon - G_\theta(x_t,t \mid c_{i-1}, y_i, r_i)\right\|^2\right],
\label{eq:self_forcing_loss}
\end{equation}
where $c_0$ is the initial context sampled from data; 
$c_{i-1}$ is the rollout context at step $i$ sampled from $p_\theta(\cdot\mid c_0)$; $x^*_{0,i}$ is the ground‑truth clean latent at step $i$; 
$y_i$ is the action input at step $i$; 
$r_i$ is the relative‑time embedding at step $i$.

This figure \ref{fig:method_view} illustrates our implementation method: The model predicts the next frame, feeds the prediction result back into the context, and applies the same diffusion loss at each step.

\subsubsection{Closed-Loop Planning}
In closed-loop planning, we select an action sequence by rolling the world model forward and minimizing a terminal cost. Formally, for a fixed initial context $c_0$, the real planning objective is:
\begin{equation}
J^*(a_{1:T})
\;=\;
\mathbb{E}_{x_{1:T}^* \sim p_{\text{data}}(\cdot \mid c_0,a_{1:T})}
\big[\, d(x_T^*, x_{\text{goal}})\,\big],
\label{eq:planning_objective_gt}
\end{equation}
where $x_{\text{goal}}$ is the goal observation, $x_{1:T}^*$ denotes the ground‑truth rollout sequence, $x_T^*$ is its final real observation at horizon $T$.

The predicted planning objective is:
\begin{equation}
J_\theta(a_{1:T}) \;=\; \mathbb{E}_{\hat{x}_{1:T} \sim p_\theta(\cdot \mid c_0,a_{1:T})}\big[\, d(\hat{x}_T, x_{\text{goal}})\,\big],
\label{eq:planning_objective}
\end{equation}
where $\hat{x}_{1:T}$ denotes the model‑predicted rollout sequence, and $\hat{x}_T$ is the final predicted observation at horizon $T$. The $d(\cdot,\cdot)$ is a LPIPS~\cite{zhang2018unreasonable} loss, and the expectation is over the model's rollout distribution $p_\theta$. Because $p_\theta$ is induced by the model's own predictions, any train--test mismatch directly affects $J_\theta$.

If $d$ is $L$-Lipschitz and $\varepsilon_i = \mathbb{E}\|\hat{x}_i - x_i^*\|$ denotes the expected per-step rollout error (under the model's distribution), then a standard Lipschitz propagation bound yields:
\begin{equation}
|J_\theta(a_{1:T}) - J^*(a_{1:T})| \;\le\; L \sum_{i=1}^{T} \varepsilon_i,
\label{eq:planning_error_bound}
\end{equation}
This inequality demonstrates that the planning error is bounded by the cumulative rollout error.

AR Forcing explicitly minimizes the diffusion loss under the model-induced distribution $p_\theta$, which directly reduces the per-step errors $\varepsilon_i$. Consequently, it tightens the bound in \eqref{eq:planning_error_bound} and improves the reliability of closed-loop action selection. This provides a theoretical justification for the observed improvements in planning performance when using AR Forcing.

\subsection{Why This Reduces Train--Test Mismatch}
We define the single-step diffusion loss as a function of the context $c$:
\begin{equation}
f_\theta(c)=
\mathbb{E}_{x_0^*,y,r,t,\epsilon}
\big[\|\epsilon-G_\theta(x_t,t\mid c,y,r)\|_2^2\big],
\end{equation}
where $c$ denotes the conditioning context, $x_0^*$ is the ground-truth future latent, $y$ is the action input, $r$ is additional conditioning, $t$ is the diffusion timestep, and $G_\theta$ is the denoiser.

Standard training optimizes this loss under data contexts, while inference uses rollout-generated contexts:
\begin{equation}
\mathcal{R}_{\text{data}}(\theta)=\mathbb{E}_{c\sim p_{\text{data}}}[f_\theta(c)],\quad
\mathcal{R}_{\text{roll}}(\theta)=\mathbb{E}_{c\sim p_\theta}[f_\theta(c)].
\end{equation}
The difference arises from the context distribution shift, which can be bounded using the total variation distance $\mathrm{TV}(p_\theta,p_{\text{data}})$:
\begin{equation}
\big|\mathcal{R}_{\text{roll}}(\theta)-\mathcal{R}_{\text{data}}(\theta)\big|
\le
\sup_c |f_\theta(c)|\,
\mathrm{TV}(p_\theta,p_{\text{data}}).
\end{equation}
This bound is included to illustrate the source of train--test mismatch, rather than as a novel theoretical contribution.

In practice, AR Forcing treats rollout contexts as stop-gradient inputs rather than backpropagating through the full temporal rollout. It therefore serves as a DAgger-style approximation, alternating between generating model rollouts and updating the denoiser on frozen rollout contexts. This is not exact end-to-end minimization of $\mathcal{R}_{\text{roll}}(\theta)$, but it exposes the model to inference-time context regions and improves long-horizon robustness.

\section{Experiments}
\label{sec:blind}
\subsection{Experimental Setting}

\subsubsection{Datasets}
We evaluate on four navigation datasets with diverse visual and motion characteristics: RECON~\cite{shah2021rapid}, TartanDrive~\cite{triest2022tartandrive}, SCAND~\cite{karnan2022socially}, and HuRoN~\cite{hirose2023sacson}. These datasets cover open-world navigation, off-road driving, urban navigation, and indoor sequences. We use the same train/test splits for all methods to avoid overlap between training and evaluation.

Each dataset provides pose information, from which we compute relative actions between adjacent frames. To unify the action scale, we normalize translation by the average step length of each dataset. Following prior navigation work~\cite{bar2025navigation}, we filter out backward-motion segments to keep the action space consistent with forward-moving navigation.

\subsubsection{Evaluation Metrics}
We report both trajectory-level and perceptual metrics. For navigation accuracy, we use Absolute Trajectory Error (ATE)~\cite{sturm2012evaluating} and Relative Pose Error (RPE)~\cite{sturm2012evaluating} to measure global drift and local consistency. We also report Position Error (PosErr) and Yaw Error (YawErr)~\cite{bar2025navigation} at the final timestep:
\begin{align}
\text{PosErr} &= \|p_t-p_t^*\|_2, \label{eq:pos_err}\\
\text{YawErr} &= \big|\mathrm{wrap}(\psi_t-\psi_t^*)\big|, \label{eq:yaw_err}
\end{align}
where $p_t$ and $p_t^*$ denote predicted and ground-truth 2D positions, and $\psi_t$ and $\psi_t^*$ denote predicted and ground-truth yaw angles. The function $\mathrm{wrap}(\cdot)$ maps angular differences to $(-\pi,\pi]$:
\begin{equation}
\mathrm{wrap}(\Delta\psi)=\operatorname{atan2}\big(\sin(\Delta\psi),\cos(\Delta\psi)\big).
\label{eq:wrap}
\end{equation}

For visual fidelity, we report LPIPS~\cite{zhang2018unreasonable}, DreamSim~\cite{fu2023dreamsim}, and FID~\cite{heusel2017gans}. All metrics are computed using identical evaluation scripts and hyperparameters across methods.

\subsubsection{Baselines}
GNM~\cite{shah2022gnm} is a general navigation model based on the setting of goal conditions. It is trained on RECON, TartanDrive, SCAND, and Go Stanford. GNM emphasizes unifying data from diverse robots and environments into a shared action space, and using goal‑conditioned trajectory prediction to drive navigation, demonstrating cross‑robot and cross‑scene generalization.

NoMaD~\cite{sridhar2024nomad} uses diffusion policies to generate navigation action sequences, supporting both goal‑conditioned navigation and goal‑free exploration. It is trained on a mixture of the GNM dataset and HuRoN. NoMaD unifies these two modes via a “goal mask” during training and uses a transformer‑based diffusion policy to model multimodal action plans. 

NWM~\cite{bar2025navigation} is a diffusion‑based navigation world model that uses CDiT to predict future visual observations and plans accordingly. It is trained on RECON, TartanDrive, SCAND, and HuRoN, enabling decision‑making in both seen and unseen environments through future‑frame generation and planning.

\begin{table}[t]
\centering
\caption{\textbf{Comparison with SOTA method on long-horizon prediction ability.}  Reporting LPIPS, DreamSim, and FID results on RECON, TartanDrive, SCAND, and HuRoN. Lower values show better performance for LPIPS, DreamSim, and FID.}
\renewcommand{\arraystretch}{0.9}
\setlength{\tabcolsep}{10pt}
\resizebox{\linewidth}{!}{%
\begin{tabular}{l c c c c c c}
\hline
\multirow{2}{*}{\textbf{Metric}} & \multirow{2}{*}{\textbf{Model}} & \multicolumn{5}{c}{\textbf{Horizon}} \\
\cline{3-7}
 &  & \textbf{1s} & \textbf{2s} & \textbf{4s} & \textbf{8s} & \textbf{16s} \\
\hline
\multicolumn{7}{c}{\textbf{RECON}} \\
\hline
\multirow{2}{*}{\textbf{LPIPS}} & NWM & 0.330 & 0.367 & 0.423 & 0.468 & 0.533 \\
                       & AR Forcing(ours) & \textbf{0.261} & \textbf{0.294} & \textbf{0.341} & \textbf{0.390} & \textbf{0.463} \\
\hline
\multirow{2}{*}{\textbf{DreamSim}} & NWM & 0.127 & 0.141 & 0.181 & 0.238 & 0.319 \\
                          & AR Forcing(ours) & \textbf{0.097} & \textbf{0.107} & \textbf{0.125} & \textbf{0.157} & \textbf{0.210} \\
\hline
\multirow{2}{*}{\textbf{FID}} & NWM & 54.8 & 57.8 & 63.0 & 70.4 & 77.3 \\
                     & AR Forcing(ours) & \textbf{45.8} & \textbf{49.3} & \textbf{52.9} & \textbf{58.0} & \textbf{66.0} \\
\hline
\multicolumn{7}{c}{\textbf{TartanDrive}} \\
\hline
\multirow{2}{*}{\textbf{LPIPS}} & NWM & 0.398 & 0.449 & 0.507 & 0.564 & 0.589 \\
                       & AR Forcing(ours) & \textbf{0.334} & \textbf{0.393} & \textbf{0.454} & \textbf{0.520} & \textbf{0.558} \\
\hline
\multirow{2}{*}{\textbf{DreamSim}} & NWM & 0.205 & 0.239 & 0.286 & 0.342 & 0.382 \\
                          & AR Forcing(ours) & \textbf{0.152} & \textbf{0.191} & \textbf{0.237} & \textbf{0.300} & \textbf{0.357} \\
\hline
\multirow{2}{*}{\textbf{FID}} & NWM & 50.1 & \textbf{49.7} & \textbf{53.4} & 63.5 & 71.1 \\
                     & AR Forcing(ours) & \textbf{47.0} & 50.2 & 54.0 & \textbf{62.2} & \textbf{70.0} \\
\hline
\multicolumn{7}{c}{\textbf{SCAND}} \\
\hline
\multirow{2}{*}{\textbf{LPIPS}} & NWM & 0.461 & 0.497 & 0.529 & 0.565 & 0.591 \\
                       & AR Forcing(ours) & \textbf{0.396} & \textbf{0.435} & \textbf{0.490} & \textbf{0.542} & \textbf{0.582} \\
\hline
\multirow{2}{*}{\textbf{DreamSim}} & NWM & 0.277 & 0.303 & 0.325 & 0.368 & \textbf{0.403} \\
                          & AR Forcing(ours) & \textbf{0.223} & \textbf{0.254} & \textbf{0.297} & \textbf{0.360} & 0.404 \\
\hline
\multirow{2}{*}{\textbf{FID}} & NWM & 92.8 & 93.5 & 95.8 & 100.4 & \textbf{101.2} \\
                     & AR Forcing(ours) & \textbf{80.7} & \textbf{87.4} & \textbf{92.5} & \textbf{97.4} & 104.8 \\
\hline
\multicolumn{7}{c}{\textbf{HuRoN}} \\
\hline
\multirow{2}{*}{\textbf{LPIPS}} & NWM & 0.346 & 0.385 & 0.435 & \textbf{0.460} & \textbf{0.514} \\
                       & AR Forcing(ours) & \textbf{0.270} & \textbf{0.315} & \textbf{0.380} & 0.466 & 0.560 \\
\hline
\multirow{2}{*}{\textbf{DreamSim}} & NWM & 0.186 & 0.208 & 0.249 & \textbf{0.289} & \textbf{0.330} \\
                          & AR Forcing(ours) & \textbf{0.129} & \textbf{0.149} & \textbf{0.200} & \textbf{0.289} & 0.392 \\
\hline
\multirow{2}{*}{\textbf{FID}} & NWM & 71.3 & 72.0 & 73.2 & 79.8 & \textbf{84.2} \\
                     & AR Forcing(ours) & \textbf{55.7} & \textbf{60.4} & \textbf{69.4} & \textbf{77.8} & 99.6 \\
\hline
\end{tabular}%
}
\label{tab:long_horizon}
\end{table}

\subsubsection{Implementation Details}
We use a CDiT backbone with a 4 frames context and the VAE tokenizer~\cite{blattmann2023stable}, matching the baseline NWM configuration. In the default experiment, we start from the checkpoint of the pretrained XL-sized NWM model, which is trained on RECON, TartanDrive, SCAND, and HuRoN. Next, we use the AR Forcing method to perform autoregressive prediction of the next 16 frames on the 4fps dataset. We use the AdamW~\cite{loshchilov2017decoupled} optimizer with a learning rate of \(8 \times 10^{-5}\). Due to privacy concerns, we are unable to obtain the high-resolution HuRoN integrated dataset. Therefore, for a fair comparison, we use 6 \(\times\) A100 80G GPUs separately to train the NWM and AR Forcing, with a total batch size of 96 and training for 50,000 steps to report the results.

\subsection{Long-Horizon Prediction}
\label{sec:long_horizon}

We evaluated the long-horizon quality under different training paradigms on four datasets. All methods used the same data partition and hyperparameters to achieve fair comparisons. We let both methods predict the future trajectory for 16 seconds in an autoregressive manner.

Table \ref{tab:long_horizon} reports the evaluation results for 1 second, 2 seconds, 4 seconds, 8 seconds, and 16 seconds under 4fps. It shows that AR Forcing can improve the stability of long-horizon prediction, while the baseline scheme shows a significant performance decline as the prediction range expands.
\begin{table}[t]
\centering
\renewcommand{\arraystretch}{0.9}
\setlength{\tabcolsep}{15pt}
\caption{\textbf{Comparison with SOTA Methods on Goal-Conditioned Visual Navigation for 2 seconds.} NWM + NoMaD ($\times$N) indicates that NoMaD generates N candidate navigation trajectories firstly, and then NWM is used to simulate and evaluate these trajectories, and the optimal one is selected for execution.}
\resizebox{\linewidth}{!}{
\begin{tabular}{l|cccccccc}
\hline
\textbf{model} &
\textbf{ATE↓} & \textbf{RTE↓}
& \textbf{PosErr↓} & \textbf{YawErr↓}\\
\hline
GNM                     & 1.87 & 0.73 & - & - \\
NoMaD                   & 1.93 & 0.52 & - & - \\
NWM + NoMaD ($\times$16) & 1.83  & 0.50 & - & - \\
NWM + NoMaD ($\times$32) & 1.78  & 0.48 & - & - \\
NWM                     & 1.38 & 0.35& 1.85 & \textbf{0.51} \\
\hline
AR Forcing (ours)              & \textbf{1.22} & \textbf{0.34} & \textbf{1.66} & 0.53 \\
\hline
\end{tabular}%
}
\label{tab:ate_rte}
\end{table}

\begin{table*}[b]  
\centering
\renewcommand{\arraystretch}{0.9} 
\setlength{\tabcolsep}{3pt} 
\caption{\textbf{Comparison on Goal-Conditioned Visual Navigation for 4s, 8s, and 16s horizons.} Compared to NWM, AR Forcing achieves better results in nearly all cases, demonstrating better prediction accuracy.}
\resizebox{\linewidth}{!}{
\begin{tabular}{l|l|ccc|ccc|ccc|ccc}
\hline
\multirow{2}{*}{Dataset} & \multirow{2}{*}{Model} 
& \multicolumn{3}{c|}{\textbf{ATE↓}} & \multicolumn{3}{c|}{\textbf{RPE↓}} 
& \multicolumn{3}{c|}{\textbf{PosErr↓}} & \multicolumn{3}{c}{\textbf{YawErr↓}} \\
 & & 4s & 8s & 16s & 4s & 8s & 16s & 4s & 8s & 16s & 4s & 8s & 16s \\
\hline
\multirow{2}{*}{RECON} 
& NWM       & 3.25 & 8.83 & \textbf{27.27} & 0.50 & 0.75 & \textbf{1.29} & 6.45 & 19.32 & 61.42 & 0.82 & 0.70 & 2.97 \\
& AR Forcing & \textbf{1.69} & \textbf{5.80} & \textbf{27.27} & \textbf{0.35} & \textbf{0.62} & \textbf{1.29} & \textbf{3.59} & \textbf{14.08} & \textbf{61.12} & \textbf{0.48} & \textbf{0.69} & \textbf{2.61} \\
\hline
\multirow{2}{*}{HuRoN} 
& NWM       & 9.57 & 31.39 & 58.26 & 1.03 & 1.72 & 1.70 & 15.97 & 51.32 & 78.13 & \textbf{0.18} & 1.34 & \textbf{1.91} \\
& AR Forcing & \textbf{9.23} & \textbf{28.76} & \textbf{56.70} & \textbf{1.00} & \textbf{1.63} & \textbf{1.67} & \textbf{15.39} & \textbf{48.23} & \textbf{75.44} & 0.47 & \textbf{1.25} & 2.87 \\
\hline
\multirow{2}{*}{SCAND} 
& NWM       & 6.68 & 18.68 & 24.61 & 0.69 & 1.05 & 0.70 & 10.85 & 32.20 & 42.49 & 0.30 & 1.06 & \textbf{0.21} \\
& AR Forcing & \textbf{4.81} & \textbf{11.33} & \textbf{18.38} & \textbf{0.51} & \textbf{0.63} & \textbf{0.54} & \textbf{7.74} & \textbf{18.40} & \textbf{31.59} & \textbf{0.09} & \textbf{0.70} & 0.54 \\
\hline
\multirow{2}{*}{TartanDrive} 
& NWM       & 15.95 & 34.85 & 72.75 & 1.78 & 1.94 & 2.05 & 27.69 & 60.26 & 124.09 & 1.05 & 0.86 & \textbf{0.30} \\
& AR Forcing & \textbf{14.49} & \textbf{32.00} & \textbf{60.61} & \textbf{1.64} & \textbf{1.75} & \textbf{1.70} & \textbf{25.41} & \textbf{54.19} & \textbf{100.62} & \textbf{0.04} & \textbf{0.29} & 0.41 \\
\hline
\end{tabular}
}
\label{tab:A1_metrics}
\end{table*}

\begin{table}[h]
\centering
\caption{\textbf{Comparison between Self Forcing and AR Forcing at 16s.} Compared with Self Forcing, AR Forcing achieves lower errors in most metrics, suggesting that preserving the standard conditional diffusion objective better supports action-conditioned long-horizon planning.}
\renewcommand{\arraystretch}{0.9} 
\setlength{\tabcolsep}{15pt}
\resizebox{\linewidth}{!}{
\begin{tabular}{l|l|cccc}
\hline
\multirow{2}{*}[0.5em]{Dataset} & \multirow{2}{*}[0.5em]{Model} & \textbf{ATE↓} & \textbf{RPE↓} & \textbf{PosErr↓} & \textbf{YawErr↓} \\
\hline
\multirow{2}{*}[0.5em]{RECON} & Self Forcing & 36.64 & 1.44 & 73.06 & 2.70 \\
& AR Forcing & \textbf{27.27} & \textbf{1.29} & \textbf{61.12} & \textbf{2.61} \\
\hline
\multirow{2}{*}[0.5em]{HuRoN} & Self Forcing & 62.65 & 1.79 & 85.68 & \textbf{0.67} \\
& AR Forcing & \textbf{56.70} & \textbf{1.67} & \textbf{75.44} & 2.87 \\
\hline
\multirow{2}{*}[0.5em]{SCAND} & Self Forcing & 30.13 & 0.84 & 52.09 & 1.62 \\
& AR Forcing & \textbf{18.38} & \textbf{0.54} & \textbf{31.59} & \textbf{0.54} \\
\hline
\multirow{2}{*}[0.5em]{TartanDrive} & Self Forcing & 82.05 & 2.28 & 139.89 & 1.15 \\
& AR Forcing & \textbf{60.61} & \textbf{1.70} & \textbf{100.62} & \textbf{0.41} \\
\hline
\end{tabular}%
}
\label{tab:16s_metrics}
\end{table}

\begin{figure}[t]
    \centering
    \includegraphics[width=1.0\linewidth]{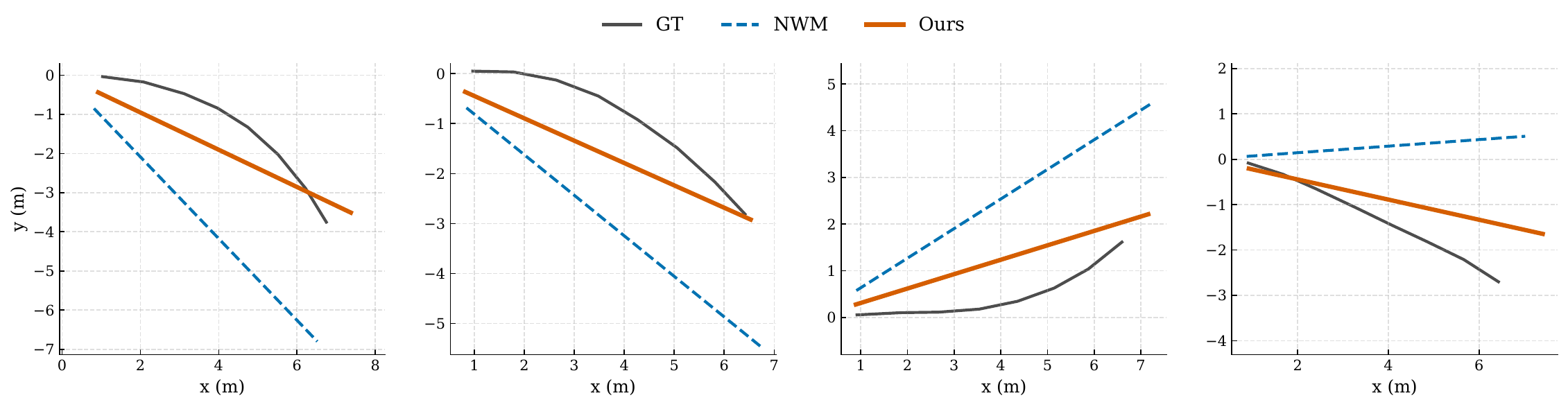}
    \caption{\textbf{Qualitative comparison of predicted trajectories.} Compared to the baseline NWM, AR Forcing (ours) produces paths that are significantly more aligned with the Ground Truth (GT), especially in complex maneuvering scenarios.}
    \label{fig:comp_traj}
\end{figure}

\subsection{Trajectory Planning}
\label{sec:trajectory_planning}
Next, we will evaluate the performance of the closed-loop planning. In this evaluation, the planner will select a set of candidate action sequences, then use the world model to simulate each candidate action, and select the action sequence with the lowest predicted cost. Specifically, using an A100 80G, with an evaluation batch size of 1, a total of 8 frames is rolled out, 120 candidate trajectories are generated for each evaluation sample, and 5 of them are selected for updating the action distribution. Finally, the evaluation is repeated 3 times and the average is taken for the result report.

Table~\ref{tab:ate_rte} reports 2s short-horizon planning results, where AR Forcing consistently improves over the baseline. Table~\ref{tab:A1_metrics} extends the evaluation to 4s, 8s, and 16s, showing lower trajectory errors and more stable relative pose estimates under long rollouts. Figure~\ref{fig:comp_traj} further visualizes that AR Forcing produces more accurate planned trajectories.

Table~\ref{tab:16s_metrics} compares AR Forcing with Self Forcing at 16s. We analyze that since robot navigation depends on action-conditioned transitions \(p(x_{t+1:t+H} \mid c, a_{1:H})\) and reliable action ranking, the standard conditional diffusion loss provides stronger action-to-outcome supervision than distribution-matching objectives mainly designed for visual fidelity.

\begin{figure}[t]
    \centering
    \includegraphics[width=1.0\linewidth]{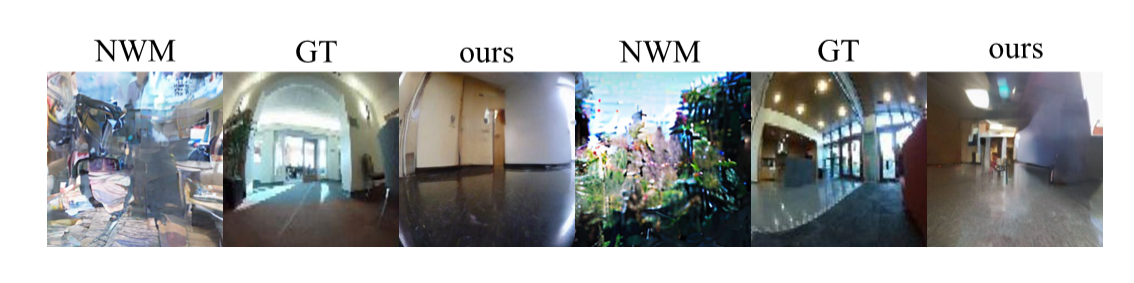}
    \caption{\textbf{Visual Qualitative Comparisons on Unknown Environment Go Stanford at 16s.} In the 16s zero-shot long-horizon generation, the drift of NWM is severe, while AR Forcing (ours) can maintain the stability of the generation.}
    \label{fig:comp_go}
\end{figure}

\begin{figure}[t]
    \centering
    \includegraphics[width=1.0\linewidth]{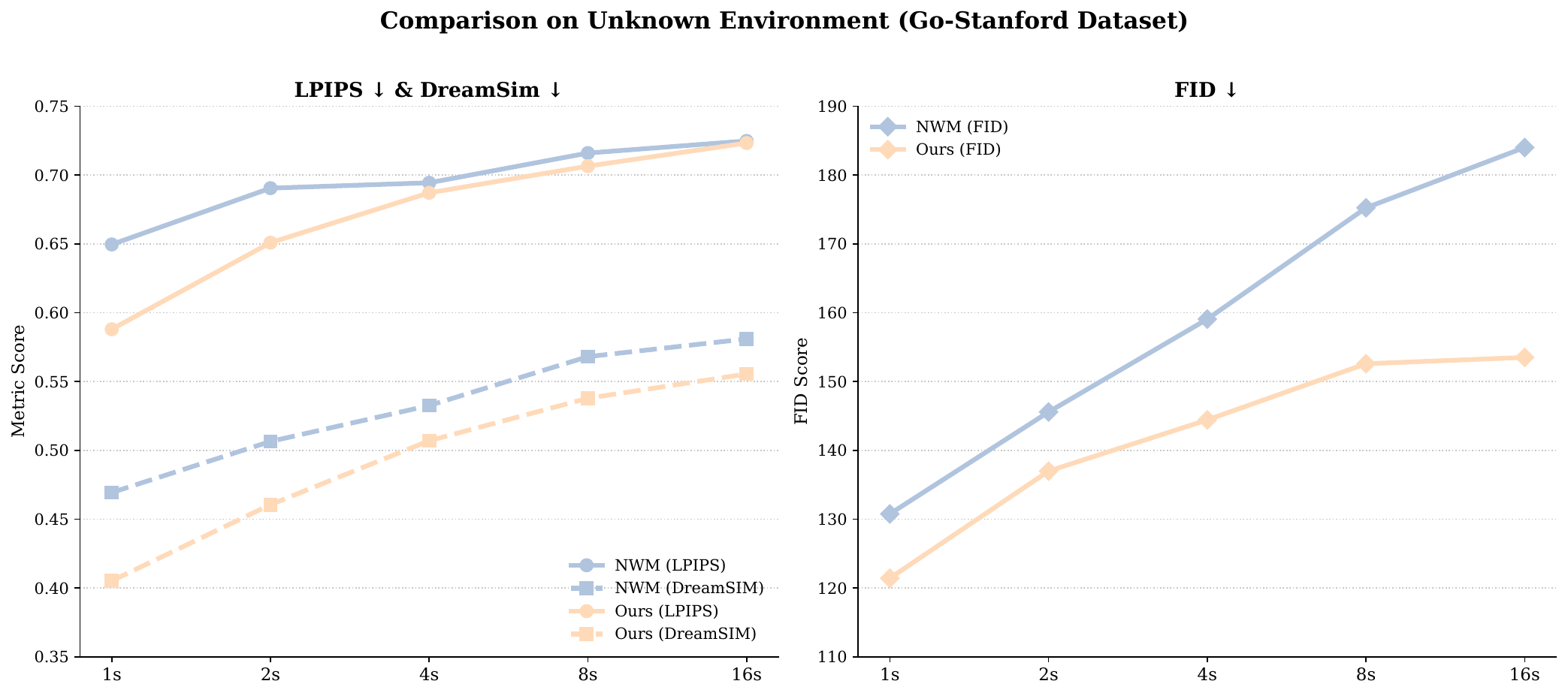}
    \caption{\textbf{Long-horizon Prediction Stability on Unknown Environment Go Stanford.} (Left) Our method achieves lower LPIPS and DreamSim scores, indicating higher temporal consistency. (Right) The flatter FID curve of AR Forcing (ours) demonstrates its effectiveness in mitigating error accumulation compared to NWM.}
    \label{fig:go_stanford_view}
\end{figure}

\begin{table}[htbp]
\centering
\caption{\textbf{Eval-loss comparison between from-scratch and pretrained initialization.} Similar convergence suggests that AR Forcing is not merely a finetuning effect.}
\renewcommand{\arraystretch}{0.9}
\setlength{\tabcolsep}{4pt}
\begin{tabular}{l|cccc}
\hline
\textbf{Training Setting / Steps} & \textbf{10k} & \textbf{20k} & \textbf{30k} & \textbf{40k} \\
\hline
From scratch & 0.8344 & 0.7071 & 0.3606 & 0.2879 \\
From pretrained & 0.8321 & 0.6944 & 0.3316 & 0.2847 \\
\hline
\end{tabular}
\label{tab:eval_loss}
\end{table}

\begin{table}[!ht]
\centering
\renewcommand{\arraystretch}{0.9}
\setlength{\tabcolsep}{15pt}

\caption{\textbf{Impact of 4, 8, and 16 frames of autoregressive prediction trajectory length on long-horizon prediction ability.} It can be observed that the longer the length of the autoregressive prediction trajectory length, the stronger the model's long-horizon prediction ability.}

\resizebox{\linewidth}{!}{%
\begin{tabular}{l c c c c c c}
\hline
\multirow{2}{*}{\textbf{Metric}} & \multirow{2}{*}{\textbf{len\_traj\_pred}} & \multicolumn{5}{c}{\textbf{Horizon}} \\
\cline{3-7}
 &  & \textbf{1s} & \textbf{2s} & \textbf{4s} & \textbf{8s} & \textbf{16s} \\
\hline
\multicolumn{7}{c}{\textbf{RECON}} \\
\hline
\multirow{3}{*}{\textbf{LPIPS}} 
& 4 & 0.278 & 0.329 & 0.401 & 0.469 & 0.534 \\
& 8 & 0.270 & 0.308 & 0.357 & 0.419 & 0.491 \\
& 16 & \textbf{0.261} & \textbf{0.294} & \textbf{0.341} & \textbf{0.390} & \textbf{0.463} \\
\hline
\multirow{3}{*}{\textbf{DreamSim}} 
& 4 & 0.106 & 0.127 & 0.173 & 0.236 & 0.306 \\
& 8 & 0.100 & 0.113 & 0.139 & 0.175 & 0.234 \\
& 16 & \textbf{0.097} & \textbf{0.107} & \textbf{0.125} & \textbf{0.157} & \textbf{0.210} \\
\hline
\multirow{3}{*}{\textbf{FID}} 
& 4 & 50.1 & 55.8 & 62.4 & 74.7 & 78.0 \\
& 8 & 47.2 & 51.2 & 54.7 & 62.3 & 69.5 \\
& 16 & \textbf{45.8} & \textbf{49.3} & \textbf{52.9} & \textbf{58.0} & \textbf{66.0} \\
\hline
\multicolumn{7}{c}{\textbf{TartanDrive}} \\
\hline
\multirow{3}{*}{\textbf{LPIPS}} 
& 4 & 0.344 & 0.412 & 0.492 & 0.573 & 0.621 \\
& 8 & \textbf{0.331} & \textbf{0.391} & 0.461 & 0.530 & 0.570 \\
& 16 & 0.334 & 0.393 & \textbf{0.454} & \textbf{0.520} & \textbf{0.558} \\
\hline
\multirow{3}{*}{\textbf{DreamSim}} 
& 4 & 0.169 & 0.217 & 0.283 & 0.375 & 0.468 \\
& 8 & 0.153 & 0.196 & 0.246 & 0.324 & 0.389 \\
& 16 & \textbf{0.152} & \textbf{0.191} & \textbf{0.237} & \textbf{0.300} & \textbf{0.357} \\
\hline
\multirow{3}{*}{\textbf{FID}} 
& 4 & 48.9 & 52.8 & 61.9 & 71.3 & 95.0 \\
& 8 & \textbf{46.1} & 50.4 & \textbf{53.9} & 62.8 & 70.7 \\
& 16 & 47.0 & \textbf{50.2} & 54.0 & \textbf{62.2} & \textbf{70.0} \\
\hline
\multicolumn{7}{c}{\textbf{SCAND}} \\
\hline
\multirow{3}{*}{\textbf{LPIPS}} 
& 4 & \textbf{0.387} & 0.441 & 0.498 & 0.555 & 0.622 \\
& 8 & 0.396 & 0.443 & 0.492 & 0.545 & 0.599 \\
& 16 & 0.396 & \textbf{0.435} & \textbf{0.490} & \textbf{0.542} & \textbf{0.582} \\
\hline
\multirow{3}{*}{\textbf{DreamSim}} 
& 4 & 0.235 & 0.269 & 0.326 & 0.406 & 0.508 \\
& 8 & 0.225 & 0.261 & 0.305 & 0.369 & 0.456 \\
& 16 & \textbf{0.223} & \textbf{0.254} & \textbf{0.297} & \textbf{0.360} & \textbf{0.404} \\
\hline
\multirow{3}{*}{\textbf{FID}} 
& 4 & 86.6 & 94.5 & 103.7 & 113.0 & 134.7 \\
& 8 & 84.0 & 89.6 & 95.2 & 102.6 & 111.6 \\
& 16 & \textbf{80.7} & \textbf{87.4} & \textbf{92.5} & \textbf{97.4} & \textbf{104.8} \\
\hline
\multicolumn{7}{c}{\textbf{HuRoN}} \\
\hline
\multirow{3}{*}{\textbf{LPIPS}} 
& 4 & 0.311 & 0.415 & 0.558 & 0.610 & 0.638 \\
& 8 & 0.272 & 0.338 & 0.460 & 0.580 & 0.613 \\
& 16 & \textbf{0.270} & \textbf{0.315} & \textbf{0.380} & \textbf{0.466} & \textbf{0.560} \\
\hline
\multirow{3}{*}{\textbf{DreamSim}} 
& 4 & 0.165 & 0.231 & 0.405 & 0.489 & 0.513 \\
& 8 & 0.139 & 0.164 & 0.264 & 0.451 & 0.493 \\
& 16 & \textbf{0.129} & \textbf{0.149} & \textbf{0.200} & \textbf{0.289} & \textbf{0.392} \\
\hline
\multirow{3}{*}{\textbf{FID}} 
& 4 & 67.8 & 86.3 & 124.5 & 154.1 & 145.0 \\
& 8 & 59.5 & 68.0 & 85.4 & 128.0 & 136.3 \\
& 16 & \textbf{55.7} & \textbf{60.4} & \textbf{69.4} & \textbf{77.8} & \textbf{99.6} \\
\hline
\end{tabular}%
}
\label{tab:ablation}
\end{table}

\subsection{Zero-Shot Generalization in the Unknown Environment}
\label{sec:unknown_env}
We evaluate zero-shot generalization on the unseen Go Stanford dataset without fine-tuning. Figures~\ref{fig:go_stanford_view} and~\ref{fig:comp_go} show that AR Forcing produces more stable long-horizon predictions than the NWM baseline, with smaller accumulated deviations and fewer failure cases.

This improvement comes from training on model-generated contexts. Compared with the baseline, which may drift into out-of-distribution states in unseen environments, AR Forcing learns to recover from its own prediction errors and therefore reduces long-horizon drift.

\subsection{Ablation Studies}
\label{sec:ablation}

\subsubsection{Effect of Pretraining Initialization}
We test whether AR Forcing depends on pretrained NWM initialization. Under the same training steps, from-scratch and pretrained models converge similarly as shown in Table~\ref{tab:eval_loss}, suggesting that AR Forcing is not merely a finetuning effect.

\subsubsection{Training Cost}
Due to rollout-context generation, AR Forcing has a 7.67$\times$ training-only overhead, reducing throughput from 0.69 to 0.09 samples/sec without adding parameters or changing the architecture.

\subsubsection{Autoregressive prediction trajectory length}
During training, we vary the number of autoregressive steps while keeping the evaluation duration fixed. Changing \texttt{len\_traj\_pred} allows the model to observe rollout-generated context chains of different lengths. As shown in Table ~\ref{tab:ablation}, increasing \texttt{len\_traj\_pred} from 4 to 8 and 16 consistently improves long-horizon stability and reduces trajectory deviations. This suggests that greater exposure to autoregressive contexts during training improves robustness during testing.

\section{Conclusion}

We propose AR Forcing, a simple training paradigm for reducing train--test mismatch in diffusion world models. By exposing the model to rollout-generated contexts during training, AR Forcing improves long-horizon stability, planning performance, and zero-shot generalization without architectural modifications.



%
%
\bibliographystyle{splncs04}
\bibliography{main}

\input{appendix.tex}
\end{document}

%% file: appendix.tex
\title{Supplementary Material} 

\author{}
\institute{}
\maketitle

\section{Implementation Details}
\label{sec:appendix_arch}
\subsection{Architecture Details}
We encode all input frames with the Stable Diffusion VAE (sd-vae-ft-ema) and scale latents by 0.18215.
Images are center-cropped and resized to $224\times224$, producing latent tensors of size $28\times28$ with 4 channels.
All training and evaluation operate in latent space; decoding to pixel space is only used for visualization and perceptual metrics.

We use a CDiT-XL/2 backbone as the denoiser. The model takes a context of latent frames and predicts a future
latent at each step. Conditioning inputs include the context latents, action tokens, and a relative-time
embedding $r$ that encodes the temporal offset of the prediction target. The transformer follows the DiT design,
with adaptive LayerNorm modulation and learned timestep embeddings. We train the model with learned variance $\sigma$ and apply EMA to all parameters for evaluation.

\subsection{Training Hyperparameter}
We train AR Forcing using the same optimizer and backbone settings as the baseline NWM, and only change the training schedule to enable autoregressive context updates. Table~\ref{tab:hp} shows the default hyperparameters used in our training process.

\begin{table}[h]
\centering
\begin{tabular}{l c}
\toprule
Training Hyperparameter & Value \\
\midrule
Context frames $K$ & 4 \\
Prediction length $T$ & 16 \\
Diffusion timesteps $T_{\text{diff}}$ & 1000 \\
Batch size & 16 \\
Total batch size & 96 \\
Optimizer & AdamW \\
Learning rate & $8\times10^{-5}$ \\
Grad clip & 10.0 \\
Learned $\sigma$ & True \\
\bottomrule
\end{tabular}
\caption{Training hyperparameters for AR Forcing.}
\label{tab:hp}
\end{table}

\section{Evaluation Details}

\subsection{Dataset Alignment via Step Normalization}
To unify action spaces across robots with different physical scales, we normalize displacement
by the dataset-specific average waypoint spacing (in meters). 
For each dataset $d$, let $\Delta_d$ denote the mean per-step distance. We transform raw displacements as:
\begin{equation}
\tilde{\mathbf{p}}_t = \frac{\mathbf{p}_t}{\Delta_d},
\end{equation}
where $\mathbf{p}_t$ denotes the raw displacement vector at step $t$ (in meters), and $\tilde{\mathbf{p}}_t$ is the corresponding normalized displacement used by the model, so that a unit step corresponds to the same approximate physical motion across datasets.
This normalization is applied consistently during training and evaluation, ensuring
that the model sees actions in a shared normalized coordinate system.

\subsection{Evaluation Modes}
Each dataset split provides a list of trajectory names. For reproducibility, we also provide precomputed evaluation indices.
Each dataset has a training split and test splits for visual evaluation and planning evaluation.

Visual evaluation is conducted with a 64-step prediction horizon and a context of 4 frames. The distance range spans $-64$ to $64$, so evaluation covers both backward and forward temporal offsets rather than a narrow forward-only window. We evaluate on the test splits of RECON, TartanDrive, SCAND, HuRoN, and Go Stanford. For each observation, we sample 4 goal conditions, which provides diversity in target directions and improves the stability of averaged metrics.

Planning is evaluated with a shorter horizon of 8 steps, still conditioned on 4 context frames, and uses a stride of 8 frames between decisions. The planning distance category is fixed to 8, which enforces a consistent step length in the planning loop and avoids mixing different step scales within a single evaluation run.

\section{More Qualitative Results}
We further provide long‑horizon qualitative comparisons by visualizing 16 seconds rollouts across multiple datasets. For each sequence, we show three synchronized columns: our method, the ground‑truth and NWM, to emphasize temporal coherence, scene consistency, and accumulation of drift over time. 

\begin{figure}[t]
    \centering
    \includegraphics[width=0.9\linewidth]{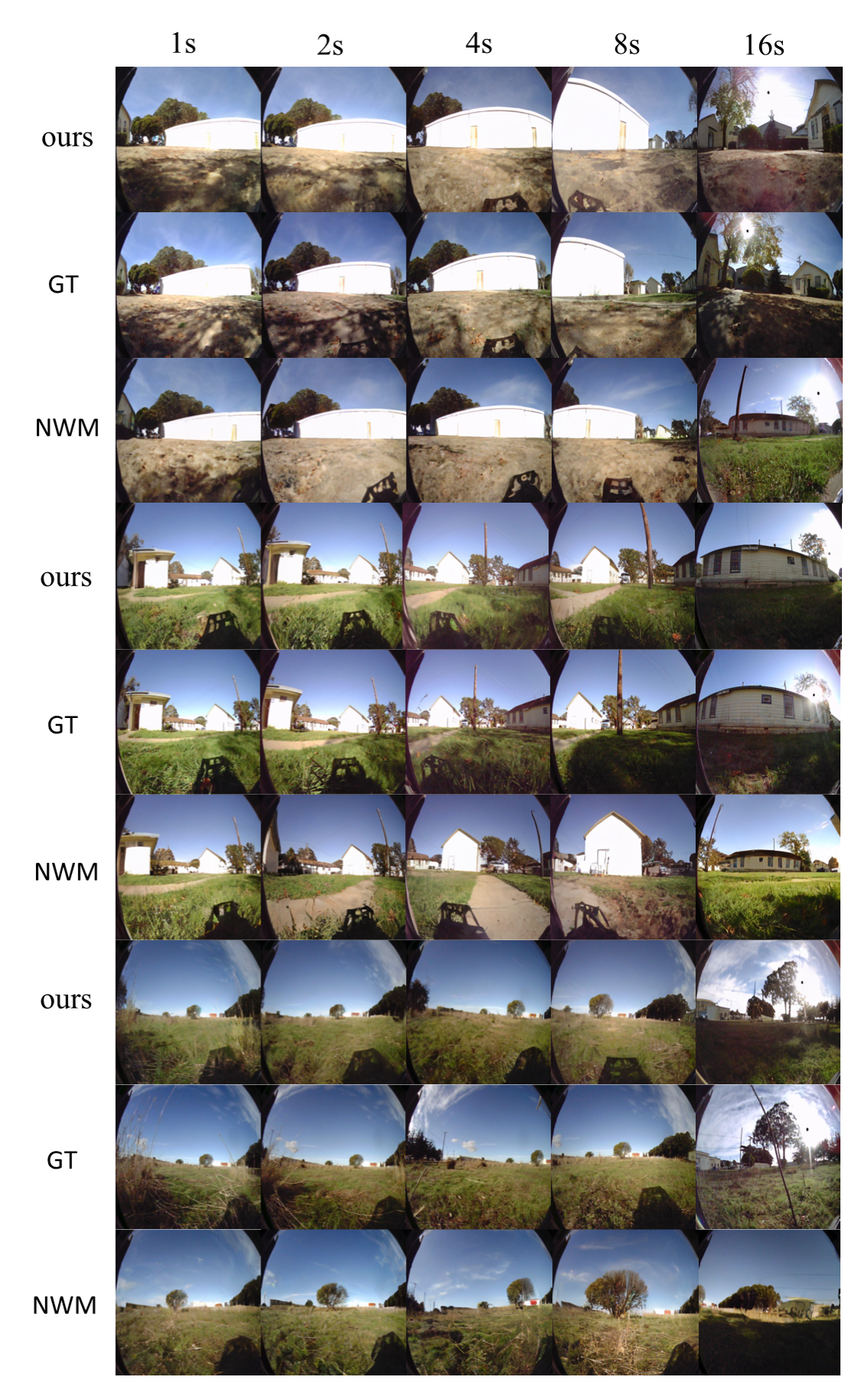}
    \caption{\textbf{Qualitative 16‑second rollouts on RECON.} AR Forcing and NWM predict the scene for the next 16 seconds at 4FPS.}
    \label{fig:recon}
\end{figure}

\begin{figure}[t]
    \centering
    \includegraphics[width=0.9\linewidth]{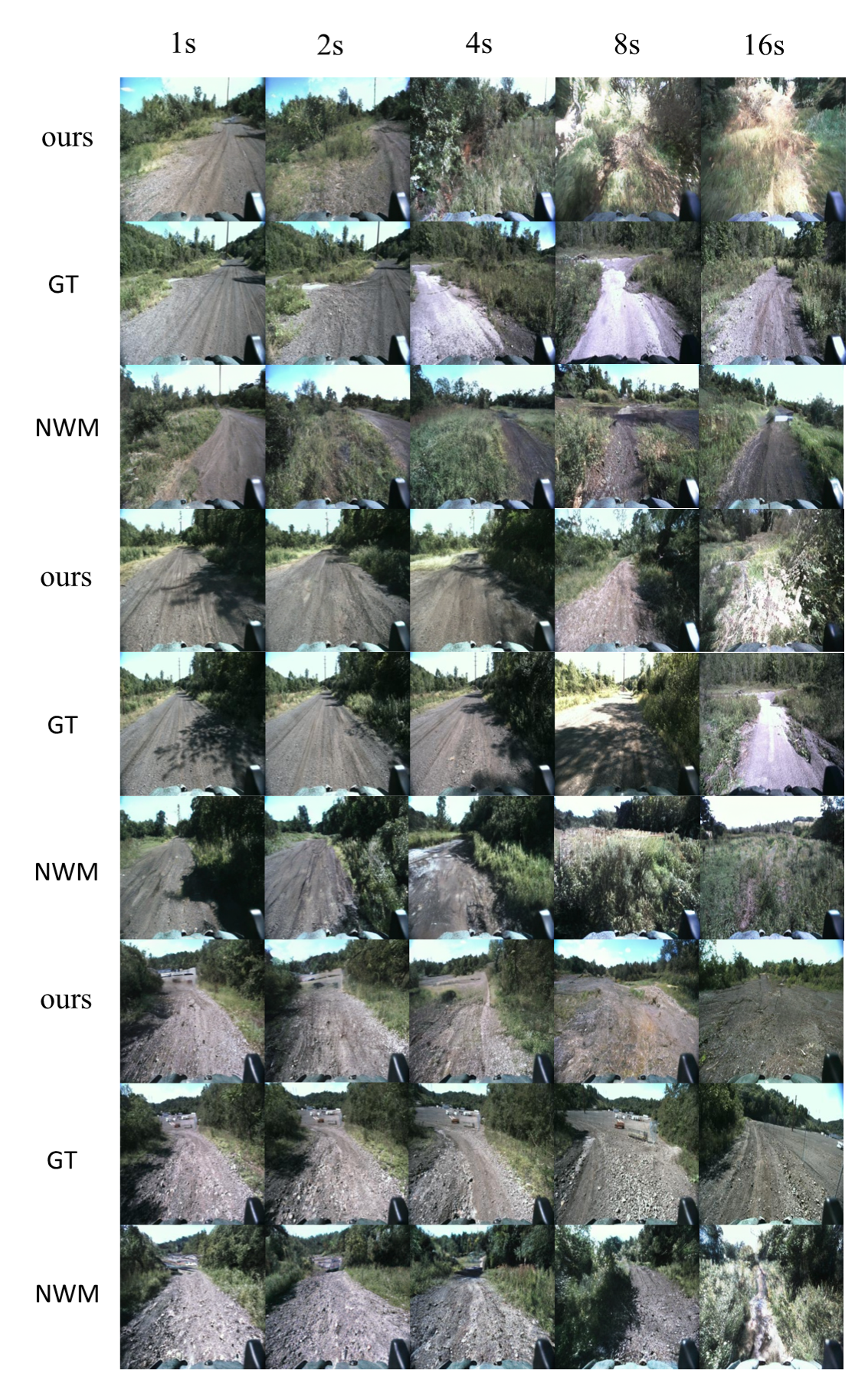}
    \caption{\textbf{Qualitative 16‑second rollouts on TartanDrive.} AR Forcing and NWM predict the scene for the next 16 seconds at 4FPS.}
    \label{fig:tartan}
\end{figure}

\begin{figure}[t]
    \centering
    \includegraphics[width=0.9\linewidth]{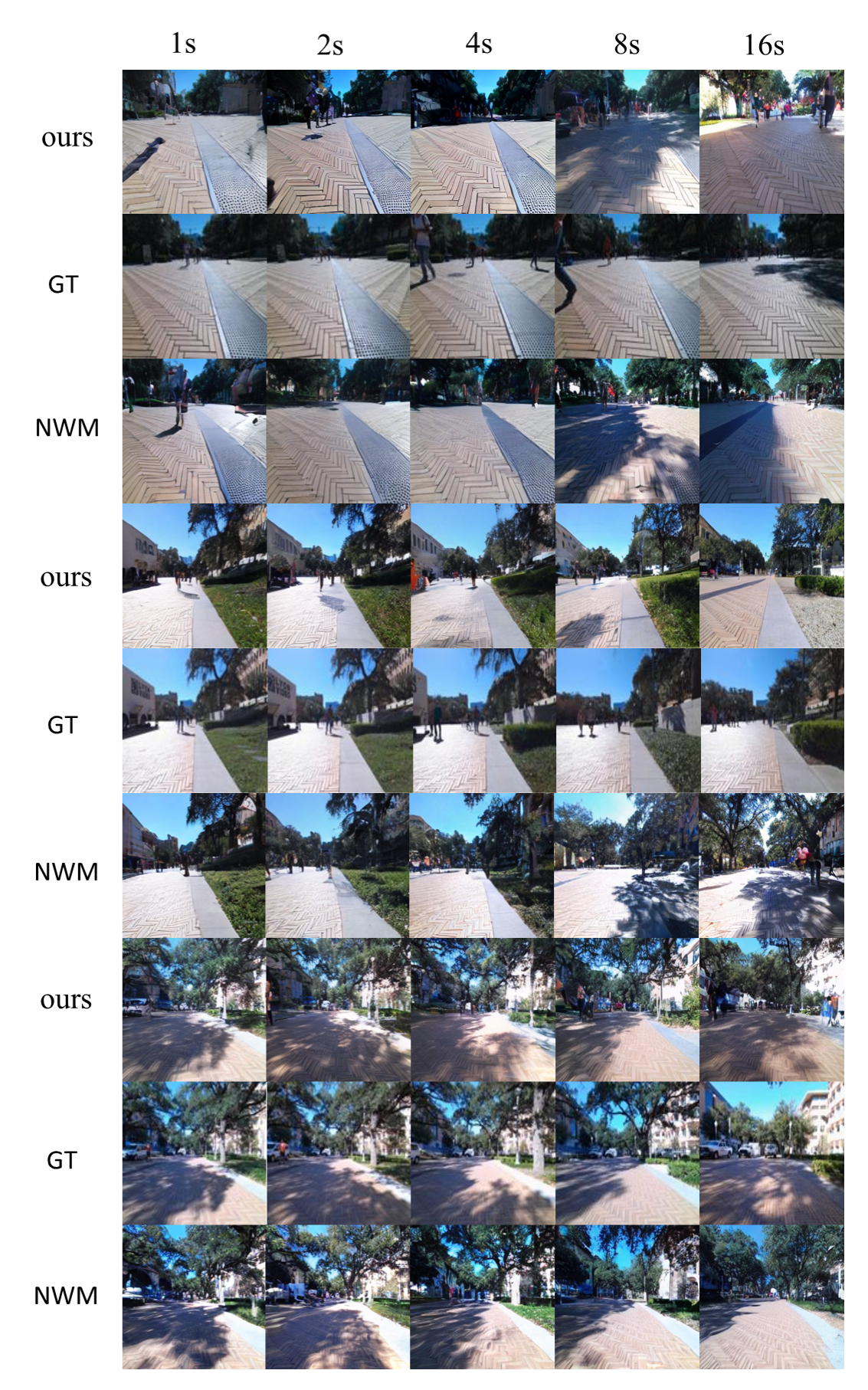}
    \caption{\textbf{Qualitative 16‑second rollouts on SCAND.} AR Forcing and NWM predict the scene for the next 16 seconds at 4FPS.}
    \label{fig:scand}
\end{figure}

\begin{figure}[t]
    \centering
    \includegraphics[width=0.9\linewidth]{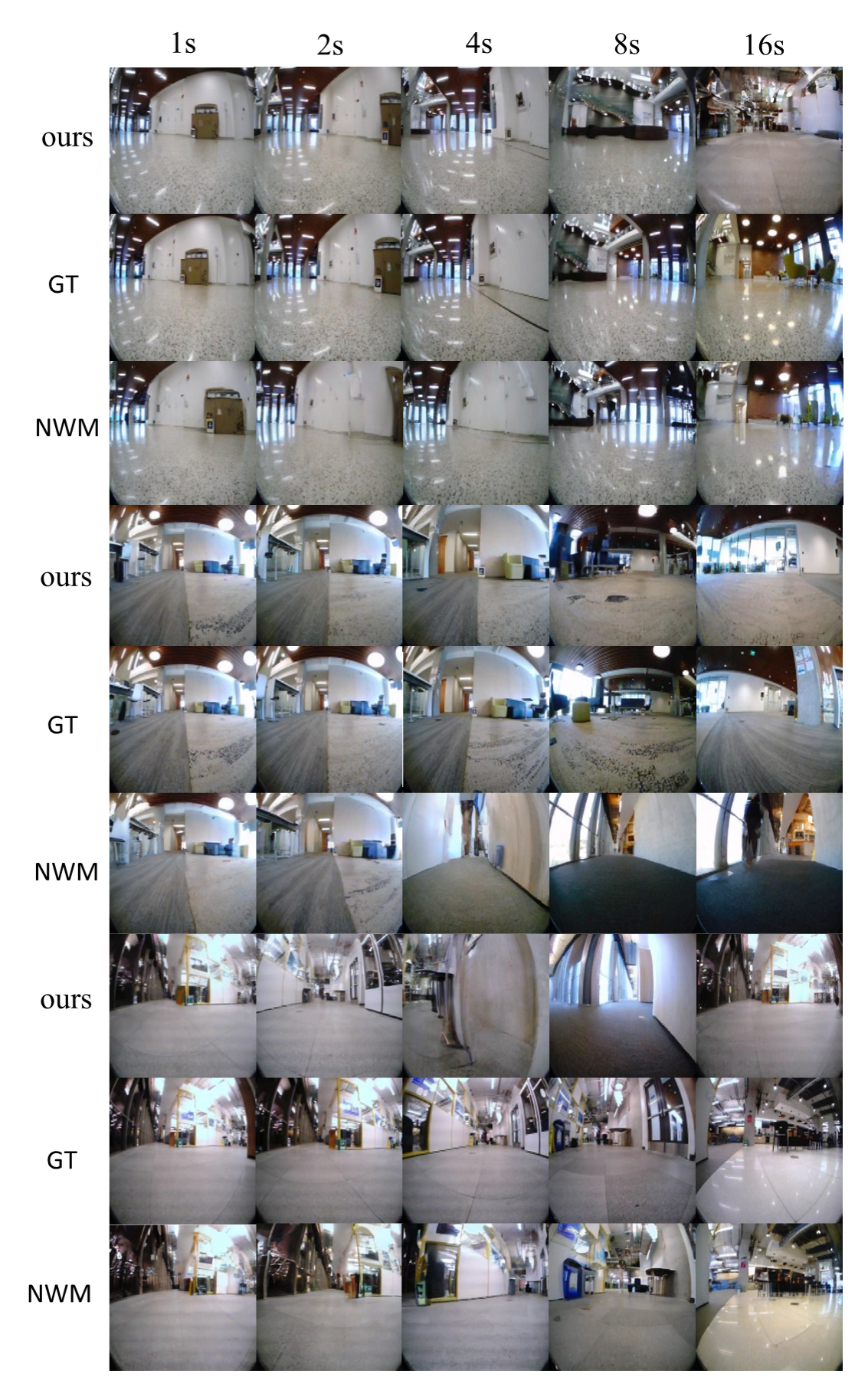}
    \caption{\textbf{Qualitative 16‑second rollouts on HuRoN.} AR Forcing and NWM predict the scene for the next 16 seconds at 4FPS.}
    \label{fig:huron}
\end{figure}

\begin{figure}[t]
    \centering
    \includegraphics[width=0.9\linewidth]{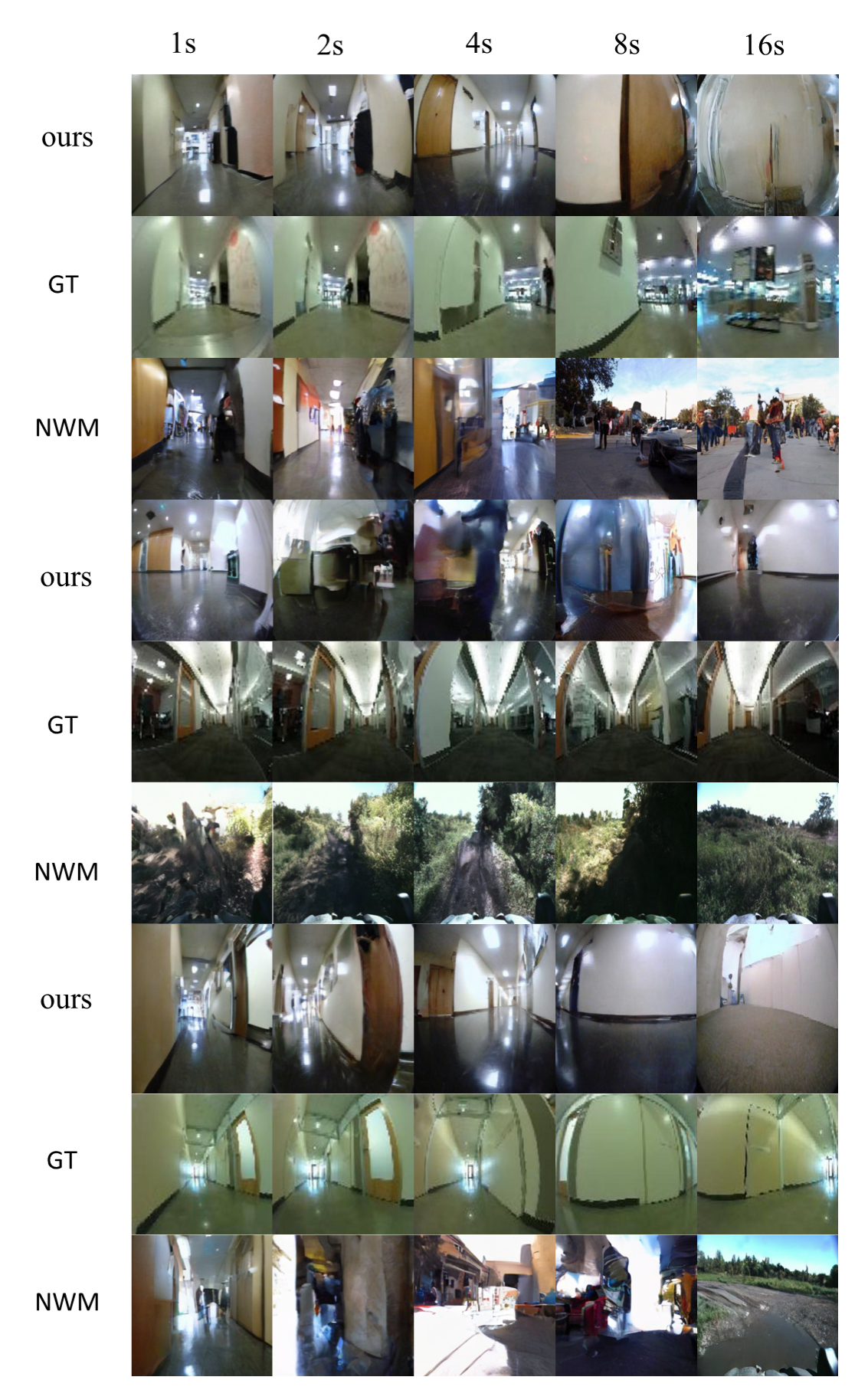}
    \caption{\textbf{Zero‑shot qualitative results on the unknown environment Go Stanford.} AR Forcing and NWM predict the scene for the next 16 seconds at 4FPS.}
    \label{fig:go}
\end{figure}

Fig~\ref{fig:recon}~\ref{fig:tartan}~\ref{fig:scand}~\ref{fig:huron} shows that qualitative results of 16‑second rollouts on RECON, TartanDrive, SCAND and HuRoN respectively. Fig~\ref{fig:go} shows zero‑shot evaluation on the unknown environment  Go Stanford.

